\documentclass[conference]{IEEEtran}
\IEEEoverridecommandlockouts
\usepackage{cite}
\usepackage{amsmath,amssymb,amsfonts}
\usepackage{algorithmic}
\usepackage{graphicx}
\usepackage{textcomp}
\usepackage{multirow}

\usepackage[colorlinks=true, linkcolor=red, citecolor=green]{hyperref}

\usepackage{xcolor}
\def\BibTeX{{\rm B\kern-.05em{\sc i\kern-.025em b}\kern-.08em
    T\kern-.1667em\lower.7ex\hbox{E}\kern-.125emX}}
\begin{document}

\title{Probablistic Restoration with Adaptive Noise Sampling for 3D Human Pose Estimation
\thanks{This work is supported by
the National Science and Technology Major Project
(Grant No.2022ZD0115904).
}
\thanks{\IEEEauthorrefmark{4} Corresponding author.}
}



\author{
\IEEEauthorblockN{Xianzhou Zeng\textsuperscript{1}, Hao Qin\textsuperscript{1}, Ming Kong\textsuperscript{1}\textsuperscript{2}, Luyuan Chen\textsuperscript{3}, Qiang Zhu\textsuperscript{1}\IEEEauthorrefmark{4}}
\IEEEauthorblockA{
\textsuperscript{1}Zhejiang University, Hangzhou, China\\
\textsuperscript{2}Hikvision Research Institute, Hangzhou, China \\
\textsuperscript{3}Beijing Information Science and Technology University, Beijing, China 
}
\{xzhouzeng, haoqin, zjukongming, zhuq\}@zju.edu.cn, chenly@bistu.edu.cn
}

\maketitle

\begin{abstract}
The accuracy and robustness of 3D human pose estimation (HPE) are limited by 2D pose detection errors and 2D to 3D ill-posed challenges, which have drawn great attention to Multi-Hypothesis HPE research. Most existing MH-HPE methods are based on generative models, which are computationally expensive and difficult to train. In this study, we propose a Probabilistic Restoration 3D Human Pose Estimation framework (PRPose) that can be integrated with any lightweight single-hypothesis model. Specifically, PRPose employs a weakly supervised approach to fit the hidden probability distribution of the 2D-to-3D lifting process in the  Single-Hypothesis HPE model and then reverse-map the distribution to the 2D pose input through an adaptive noise sampling strategy to generate reasonable multi-hypothesis samples effectively. Extensive experiments on 3D HPE benchmarks (Human3.6M and MPI-INF-3DHP) highlight the effectiveness and efficiency of PRPose.  Code is available at: \url{https://github.com/xzhouzeng/PRPose}.
\end{abstract}


\section{Introduction}

3D Human Pose Estimation (HPE) is a crucial task for restoring the 3D position of human joints from a monocular image, which has broad applications in virtual reality, action recognition and human-computer interaction, etc \cite{10.1145/3524497}. Current mainstream 3D HPE solutions follow a two-stage approach: first to estimate the 2D human pose and then predict the 3D pose based on the estimated 2D results, i.e., 2D-to-3D lifting \cite{zou2021modulated, cai2023htnet}. Among them, 2D-to-3D lifting is the pivotal stage, which presents two critical challenges: Firstly, the mapping from 2D to 3D is ill-posed, as a single 2D pose can correspond to multiple plausible 3D poses, impacting model convergence. Secondly, accurate 2D estimation is essential, as errors amplify during lifting, leading to unpredictable discrepancies in final 3D estimation outcomes \cite{martinez2017simple}.

\begin{figure}[tpb]
\centering
\includegraphics[width=1\columnwidth]{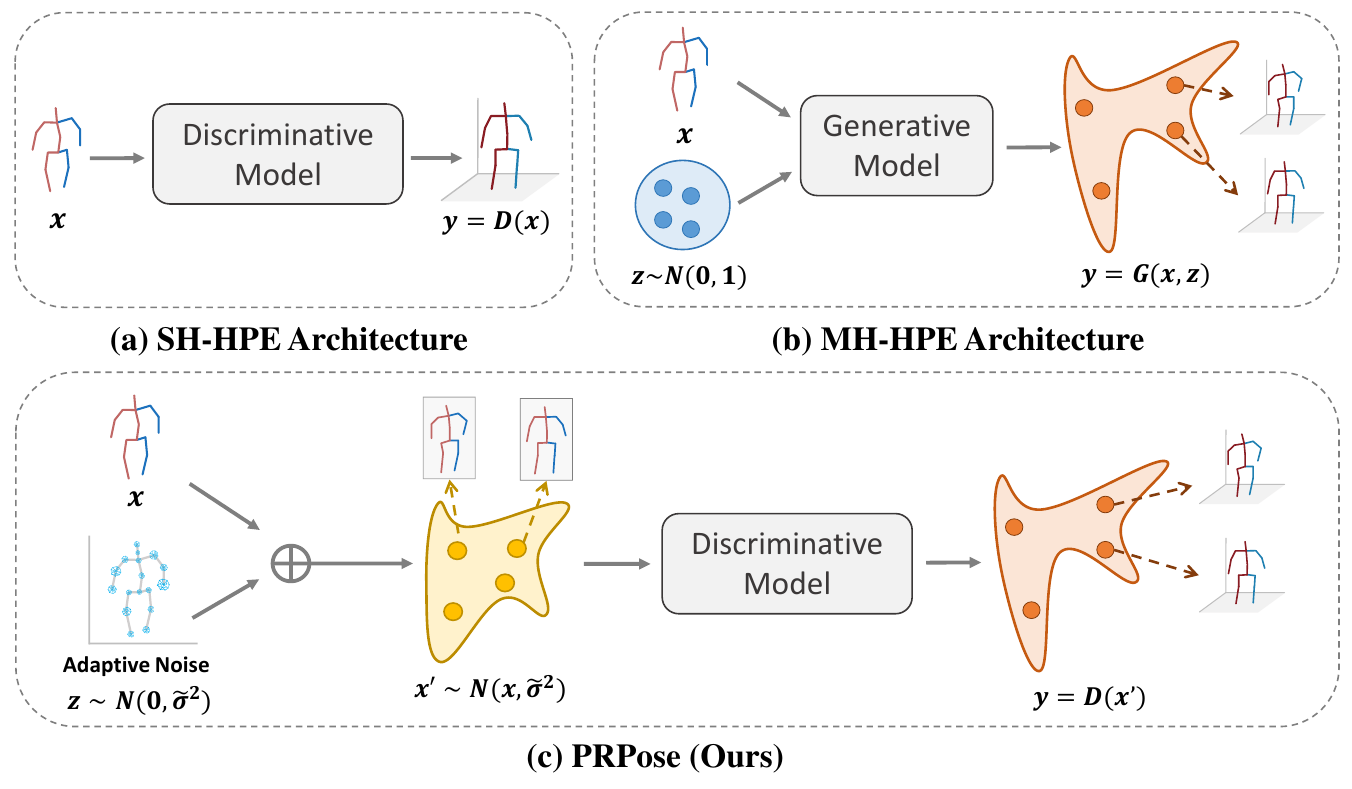} 
\caption{(a) The classical SH-HPE architecture produces a unique 3D pose based on a single input 2D pose; (b) The traditional MH-HPE architecture based on the generative model, adding random noise to generate multi-hypothesis results; (c) We propose PRPose that can be extended from any SH-HPE method, generating the input distribution through adaptive noise to reconstruct the original probabilistic modeling process of SH-HPE, and producing multi-hypothesis outputs.}
\label{fig1}
\end{figure}

Early methods for 2D-to-3D lifting estimate a unique 3D pose from a given 2D pose, named Single-Hypothesis Human Pose Estimation (SH-HPE), as illustrated in Figure \ref{fig1}.a, which primarily employ discriminative models, such as regression networks. These methods have high computational efficiency but suffer from the ill-posed nature of the 2D-to-3D mapping and inaccurate 2D pose estimation \cite{martinez2017simple}. Recent research \cite{sharma2019monocular} addresses these issues through Multi-Hypothesis Human Pose Estimation (MH-HPE), which leverages probabilistic modeling to accommodate estimation errors stemming from imprecise 2D pose detection or ill-posed mapping, generating multiple distinct 3D pose outputs aligned with the input 2D pose.

Most MH-HPE methods adopt generative models (e.g., Diffusion Model) to enhance 2D-3D mapping for 3D human pose estimation, as shown in Figure \ref{fig1}.b. These methods introduce fixed-distribution noise into the generative model to diversify the output \cite{wehrbein2021probabilistic, ci2023gfpose}. However, such processes are computationally demanding and challenging to converge. Moreover, existing methods employ uniform perturbations across all joints, disregarding variations in pose and joint complexities. Consequently, joints with higher uncertainty, such as those on occluded limbs, may suffer from insufficient hypothesis generation, while easily estimable joints, such as trunk joints, could experience unnecessary perturbations, leading to excessive disturbances and ineffective hypothesis sampling.

 Recent studies have revealed that the SH-HPE method has potential probabilistic modeling capabilities in 2D-to-3D lifting \cite{li2023poseoriented}. Inspired by that, we propose a novel PRPose method (Fig. \ref{fig1}.c) to take full advantage of the probabilistic modeling capabilities of SH-HPE models. Specifically, we propose a weakly supervised adaptive noise sampling strategy that creates pseudo-labels from the single-hypothesis model to generate variable noise scales for error-prone joints. Through a weakly supervised learning process, PRPose recovers the probabilistic modeling process of the SH-HPE model, generating an adaptive variance matrix from the 2D pose, and remapping it to create adaptive noise for the 2D input joints. Finally, the SH-HPE model generates multiple 3D pose hypotheses based on different sampled 2D poses. Compared to previous methods, the scaling noise generated by PRPose adapts to 2D pose and joint topology, enhancing the realism of 3D pose assumptions.

Extensive experiments conducted on the Human3.6M and MPI-INF-3HP datasets show the comparable performance of the PRPose to the state-of-the-art approaches with a speed enhancement of two orders of magnitude. Comprehensive ablation experiments and visualizations demonstrate the reasoning for using adaptive noise addition and the effectiveness of a weakly supervised training strategy. Our main contributions can be summarized as follows:

\begin{itemize}
 \item The proposed architecture enables simple extension of any SH-HPE model to the MH-HPE task. We preserve the lightweight and trainable advantages of the SH-HPE model, while suppressing the inherent flaws of the SH-HPE, such as 2D estimation error and ill-posed nature in the 2D-to-3D mapping process.
 \item We design a weakly supervised adaptive noise learning strategy that generates adaptive noise based on input 2D poses and joint positions. During the 2D-to-3D lifting process, we apply large noise corrections to joints with potentially large estimation errors, thereby improving the generated multi-hypothesis hit rate.
 \item Our method achieves a good balance between speed and accuracy. On the Human3.6M and MPI-INF-3DHP datasets, we achieve comparable accuracy to GFPose \cite{ci2023gfpose} (SOTA), but with over 100x faster pose estimation.
\end{itemize}

\section{Related Work}
Early works on 3D HPE mainly estimate 3D poses via a end-to-end network \cite{pavlakos2017coarse, chen2021joint, sun2017compositional, mehta2017vnect, wei2016convolutional}. However, the highly nonlinear nature of mapping from 2D images to 3D poses caused challenges of extensive search space. Benefiting from the success of 2D HPE \cite{newell2016stacked, chen2018cascaded, sun2019deep}, the 2D-to-3D lifting methods \cite{ma2021context, xu2021graph}, which infers 3D human poses from intermediate estimated 2D poses, has become a popular solution.

\subsection{SH-HPE}

Martinez \emph{et al.} \cite{martinez2017simple} first proposed a two-stage human pose estimation method that generates a unique 3D pose estimation from the 2D input, called Single-Hypothesis HPE (SH-HPE). They also proved that the error of the two-stage methods is mainly caused by the 2D pose estimation errors. Subsequent works tried to minimize the ambiguity in the 2D-to-3D lifting process \cite{zhou2021hemlets}, or propose data augmentation strategies to address the issue of limited available data and decrease the influence of data bias \cite{li2020cascaded, gong2021poseaug}. The information transmission between joints is another focus. Researchers tried to achieve the hidden-state fusion between joints with GCN \cite{zou2021modulated, li2022graphmlp} and Transformer \cite{zhao2022graformer}, so as to enhance the representation ability and benefit highly-uncertainty joints prediction.

Most SH-HPE methods employ discriminative models, which are computationally cheap and easy to train. However, limited by a single hypothesis output, SH-HPE methods still face challenges in solving the ill-posed problem of 2D to 3D lifting and 2D pose error transfer \cite{jahangiri2017generating}.

\subsection{MH-HPE}

To address the limitations of SH-HPE, Jahangiri \emph{et al.} \cite{jahangiri2017generating} first proposed the idea of multiple hypotheses estimation (MH-HPE), with the main purpose of improving the accuracy and robustness adapted to various complexity and variability of scenes. MH-HPE aims to model the 2D-3D lifting process probabilistically, generating multiple hypotheses using different parameter combinations and constraints. Finally, these hypotheses can be further screened and fused to derive the most reasonable 3D pose estimation results. 

Early MH-HPE methods proposed models with multiple prediction heads to achieve multi-hypothesis results \cite{li2019generating, oikarinen2021graphmdn}. Thanks to the excellent capability of probabilistic modeling, most recent works applied generative models on the MH-HPE task, such as conditional variational autoencoder \cite{kingma2014semi, sharma2019monocular} or normalizing flow \cite{wehrbein2021probabilistic}. With the enormous improvement in the generation capabilities of diffusion models \cite{song2021maximum}, \cite{ci2023gfpose} propose GFPose and significantly improved the state-of-the-art of MH-HPE. Recently, Shan et al. used the diffusion model to explore MH-HPE based on sequence input \cite{Shan_2023_ICCV}. Although the generative model-based approaches have natural plausibility and advantages when dealing with depth blur and 2D pose estimation errors, these methods are plagued by substantial computational complexity and training difficulty. Moreover, they primarily introduce all joints to an equaled noise sampled from a predetermined pre-learned distribution, with limited regard for disparities in joint behavior across distinct human poses.

We proposed a novel PRPose enables the direct extension of lightweight single-hypothesis models to a multi-hypothesis approach. By employing adaptive noise sampling, it effectively reconstructs the original probabilistic modeling process of the SH-HPE method, so that to retain efficiency while generating more reasonable and diverse hypotheses.

\begin{figure*}[t]
\centering
\includegraphics[width=1\textwidth]{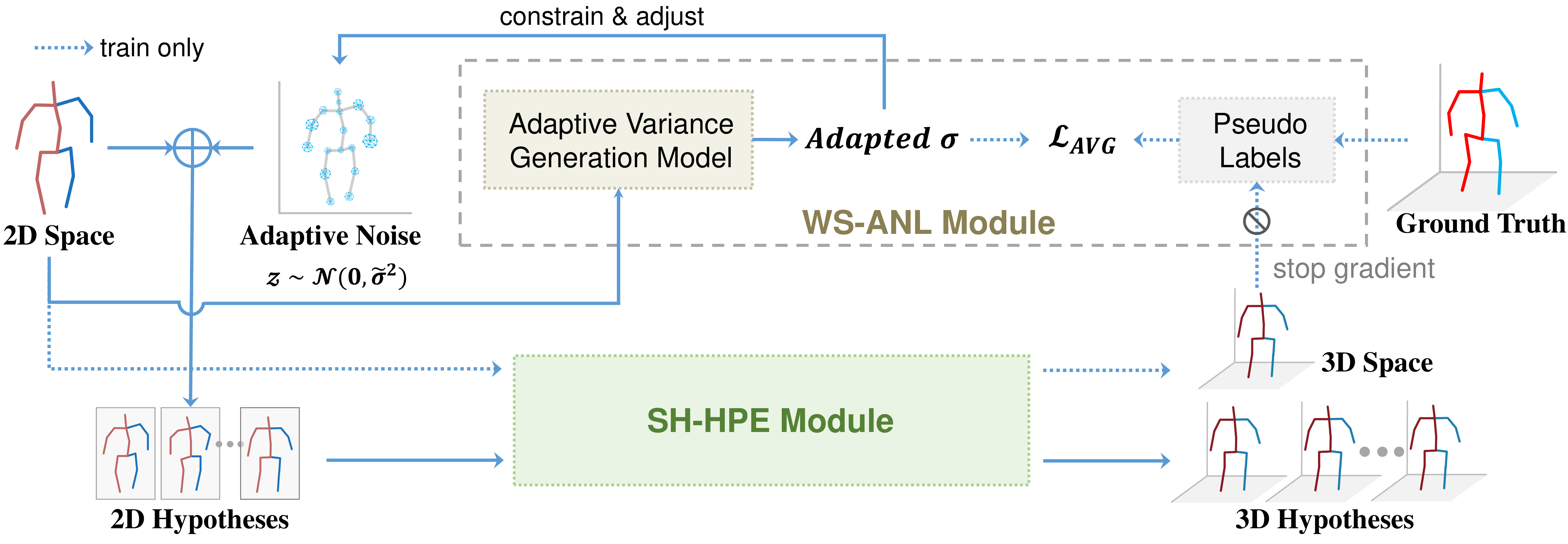} 
\caption{Our proposed architecture PRPose consists of two main modules: a) Weakly Supervised Adaptive Noise Learning (WS-ANL): completes weakly supervised training of Adaptive Variance Generation Model; b) Single-Hypothesis Human Pose Estimation (SH-HPE): superimpose adaptive noise to the original 2D pose to generate multiple 2D hypotheses; map a single 2D pose to a 3D pose by an SH-HPE model (such as HTNet) during the inference process.}
\label{fig2}
\end{figure*}

\section{Method}
\subsection{Overview}
Our work aims to extend lightweight single-hypothesis models to the multi-hypothesis method to simplify pose generation and enhance the efficiency of the 3D HPE task. Our proposed framework, shown in Figure \ref{fig2}, comprises two main modules: the Single-Hypothesis HPE (SH-HPE) and the Weakly Supervised Adaptive Noise Learning (WS-ANL). The SH-HPE module is the major task to train a discriminative single-hypothesis model, which is free to plug in existing frameworks (e.g., HTNet). The WS-ANL module is the auxiliary task to train an adaptive noise generator that is weakly supervised by pseudo-labels based on the difference between single-hypothesis prediction results and 3D pose ground truth. The inference process is performed in two steps. Firstly, we predict adaptive noise and add it to the original 2D pose to produce multiple augmented poses. Next, we input these 2D samples into a single-hypothesis module to generate multiple 3D poses.

Specifically, given the input 2D pose $x \in \mathbb{R}^{2V}$ that contains $V$ joints, our method can be expressed as:
\begin{equation}
  Y = \Phi(\Psi(x,S))
\end{equation}

\noindent where $Y =\{y_1,y_2,...,y_S | y_i\in \mathbb{R}^{3V}\}$ is the multi-hypothesis output of the model, $\Psi$ is the sample amplification step that relies on adaptive noise, and $S$ is the number of amplifications for 2D samples, $\Phi$ is a single-hypothesis model for HPE task.

Next, we will briefly explain the working mechanism of the SH-HPE and the models adopted in our experiments and then introduce the design and implementation of the adaptive noise generation module in detail.

\subsection{SH-HPE Module}

SH-HPE models are usually based on discriminative networks, and their working paradigm can be expressed as:
\begin{equation}
  y = \Phi_{\theta}(x)
\end{equation}
where $y\in \mathbb{R}^{3V}$ is a single 3D skeleton prediction, and $\theta$ is the trainable parameters in the SH-HPE model $\Phi$. Generally, the training process of $\Phi$ is supervised by the ground truth 3D skeleton:
\begin{equation}
  \mathcal{L} = |y - \hat{y}|
\end{equation}
where $\hat{y}$ is the ground truth 3D skeleton. SH-HPE often uses a model with a small amount of calculation and low training consumption (such as a regression model) as the backbone network of the framework. Our method can be adapted to any existing SH-HPE model and directly use the trained weights without minor adjustments.

Note that our framework can be seamlessly integrated with any SH-HPE model and can be customized to meet various scene-specific requirements, achieving an optimal balance of estimation accuracy and computational performance. In experiments, we compared two classic SH-HPE models HTNet \cite{cai2023htnet} and MGCN \cite{zou2021modulated}. The specific imact of different SH-HPE backbone models on the accuracy and speed of 3D HPE task is compared and analyzed.

\begin{table*}[t]
\setlength{\tabcolsep}{1.2pt}
\centering
\small
\caption{Pose estimation results on the Human3.6M dataset. We report the minMPJPE(mm) under Protocol \#1 (no rigid alignment) and Protocol \#2 (with rigid alignment). $S$ denotes the number of hypotheses. Bold and underlined highlight the optimal and suboptimal results, respectively, and the * mark highlights the optimal/suboptimal results under $S=10$.}
\resizebox{\textwidth}{!}{

\begin{tabular}{lc|ccccccccccccccc|c}
\hline
\multicolumn{2}{c|}{Protocol \#1}              & Dire. & Disc. & Eat  & Greet & Phone & Photo & Pose & Purch. & Sit  & SitD  & Smoke & Wait & WalkD & Walk & WalkT & Avg  \\ \hline
GAN \cite{li2020weakly} ($S=10$) &BMVC'20 & 62.0  & 69.7  & 64.3 & 73.6  & 75.1  & 84.8  & 68.7 & 75.0   & 81.2 & 104.3 & 70.2  & 72.0 & 75.0  & 67.0 & 69.0  & 73.9* \\
MDN \cite{li2019generating} ($S=5$) &CVPR'19  & 43.8  & 48.6  & 49.1 & 49.8  & 57.6  & 61.5  & 45.9 & 48.3   & 62.0 & 73.4  & 54.8  & 50.6 & 56.0  & 43.4 & 45.5  & 52.7 \\
CVAE \cite{sharma2019monocular} ($S=200$)  &ICCV'19   & 37.8  & 43.2  & 43.0 & 44.3  & 51.1  & 57.0  & 39.7 & 43.0   & 56.3 & 64.0  & 48.1  & 45.4 & 50.4  & 37.9 & 39.9  & 46.8 \\
GraphMDN \cite{oikarinen2021graphmdn} ($S=200$)  &IJCNN'21  & 40.0  & 43.2  & 41.0 & 43.4  & 50.0  & 53.6  & 40.1 & 41.4   & 52.6 & 67.3  & 48.1  & 44.2 & 44.9  & 39.5 & 40.2  & 46.2 \\
NF \cite{wehrbein2021probabilistic} ($S=200$) &ICCV'21   & 38.5  & 42.5  & 39.9 & 41.7  & 46.5  & 51.6  & 39.9 & 40.8   & 49.5 & 56.8  & 45.3  & 46.4 & 46.8  & 37.8 & 40.4  & 44.3 \\
GFPose \cite{ci2023gfpose} ($S=10$) &CVPR'23  & 39.9  & 44.6  & 40.2 & 41.3  & 46.7  & 53.6  & 41.9 & 40.4   & 52.1 & 67.1  & 45.7  & 42.9 & 46.1  & 36.5 & 38.0  & \underline{45.1*} \\
GFPose \cite{ci2023gfpose} ($S=200$) &CVPR'23 & 31.7  & 35.4  & 31.7 & 32.3  & 36.4  & 42.4  & 32.7 & 31.5   & 41.2 & 52.7  & 36.5  & 34.0 & 36.2  & 29.5 & 30.2  & \textbf{35.6} \\ \hline
\textbf{Ours} ($S=10$)      &         & 40.1  & 44.3  & 39.7 & 43.1  & 45.5  & 50.3  & 41.8 & 40.8   & 50.5 & 58.6  & 44.4  & 41.7 & 47.6  & 35.9 & 36.9  & \textbf{44.1*} \\
\textbf{Ours} ($S=200$)      &        & 34.5  & 38.3  & 33.9 & 37.1  & 39.8  & 43.3  & 35.6 & 35.3   & 44.3 & 51.0  & 38.9  & 36.1 & 41.7  & 31.2 & 31.7  & \underline{38.2} \\ \hline \hline
\multicolumn{2}{c|}{Protocol \#2}                & Dire. & Disc. & Eat  & Greet & Phone & Photo & Pose & Purch. & Sit  & SitD  & Smoke & Wait & WalkD & Walk & WalkT & Avg  \\ \hline
CVAE \cite{sharma2019monocular} ($S=200$)  &ICCV'19   & 30.6  & 34.6  & 35.7 & 36.4  & 41.2  & 43.6  & 31.8 & 31.5   & 46.2 & 49.7  & 39.7  & 35.8 & 39.6  & 29.7 & 32.8  & 37.3 \\
GraphMDN \cite{oikarinen2021graphmdn} ($S=200$)  &IJCNN'21  & 30.8  & 34.7  & 33.6 & 34.2  & 39.6  & 42.2  & 31.0 & 31.9   & 42.9 & 53.5  & 38.1  & 34.1 & 38.0  & 29.6 & 31.1  & 36.3 \\
NF \cite{wehrbein2021probabilistic} ($S=200$)   &ICCV'21 & 27.9  & 31.4  & 29.7 & 30.2  & 34.9  & 37.1  & 27.3 & 28.2   & 39.0 & 46.1  & 34.2  & 32.3 & 33.6  & 26.1 & 27.5  & 32.4 \\
GFPose \cite{ci2023gfpose} ($S=200$) &CVPR'23 & 26.4  & 31.5  & 27.2 & 27.4  & 30.3  & 36.1  & 26.8 & 26.0   & 38.4 & 45.8  & 31.2  & 29.2 & 32.2  & 23.1 & 25.8  & \textbf{30.5} \\ \hline
\textbf{Ours} ($S=200$)        &      & 27.7  & 31.2  & 28.6 & 30.5  & 32.2  & 35.1  & 28.2 & 27.9   & 36.9 & 41.4  & 32.2  & 28.4 & 33.4  & 24.7 & 26.1  & \underline{31.0} \\ \hline
\end{tabular}
}
\vspace*{-1em}
\label{tab1}
\end{table*}

\subsection{WS-ANL Module}
The main purpose of adding noise to the MH-HPE method is to improve algorithm robustness. A strategy we often adopt is to generate more pose hypotheses by introducing random noise in the pose generation step in order to improve the hit rate of the 3D estimation results compared to the real pose. This technique can effectively improve the search range and space of the algorithm, reduce the overlap between hypotheses, and improve the robustness, accuracy, and generalization ability of the algorithm. However, the random noise strategy does not consider the variability brought by different poses and joints. It cannot fully utilize the prior information of 2D pose and the human body model itself. Therefore, we train an adaptive variance generative model ($AVG$) with generalization ability through weakly supervised learning. The adaptive model can add customized noise according to pose and joints to achieve the best pose estimation effect:
\begin{equation}
  A^{\sigma}=A V G_{\varphi}(x)
\end{equation}
where $AVG$ represents the Adaptive Variance Generation model, $\varphi$ is its parameter, and $A^{\sigma} = \{\sigma_1,...,\sigma_V\}$ is the set of output adaptive variance.

Without relying on the human-craft labels, we employ pseudo-labels for weakly supervised training to $AVG$. We hope that the output of $AVG$ can be proportional to the magnitude of the error on each joint. Therefore, we first use the existing SH-HPE model to estimate 3D pose $y$ corresponding to the current 2D input $x$, and then calculate the difference between the 3D estimate and the 3D ground truth as a supervisory signal:
\begin{equation}
  \hat{A}^{\sigma}=\frac{D(y, \hat{y})}{C}
\end{equation}
where $\hat{A}^{\sigma} \in \mathbb{R}^{V}$ is treated as a pseudo-label to supervise the output of $AVG$, $D(\cdot,\cdot)$ represents the calculation of Euclidean distance, and $C$ is a statistical normalization constant:
\begin{equation}
  C=\frac{\sum_{p \in P} \sum_{v \in V} D\left(y_{v}^{p}, \hat{y}_{v}^{p}\right)}{P V}
\end{equation}
where $P$ is the number of training samples in the dataset. $C$ counts the average single-hypothesis estimation error of each joint of all samples in the training set. We use it as a normalization constant to stabilize the training process. Finally, calculate the loss value of this process and optimize $AVG$ by gradient descent method:
\begin{equation}
  \mathcal{L}_{A V G}=\sum_{\mathbf{x}} \frac{\sum_{v \in V}\left(A_{v}^{\sigma}-\hat{A}_{v}^{\sigma}\right)^{2}}{V}
\end{equation}

During this process, the parameters of the SH-HPE model are frozen and do not participate in the update process.

In the test phase, for a specific 2D input $x$, the Gaussian noise distribution  $z_v\sim\mathcal{N}\left(0, \tilde{\sigma_v}^2\right)$  corresponding to each joint can be obtained after constrained and adjusted:
\begin{equation}
  \tilde{\sigma_v}=\alpha *\max \left(\sigma_v,1\right)
\end{equation}
where $\alpha$ is a coefficient for adjusting the noise distribution. This operation can increase the noise variance of joints with minor errors while maintaining the noise variance of joints with significant errors unchanged, ensuring the adaptability of sampling and avoiding weakening the solution ability of the MH-HPE method to the ill-posed problem due to the weak noise addition of some joints. A new 2D sample $x^{\prime}$ is obtained by randomly sampling a noise $z_v\sim\mathcal{N}\left(0, \tilde{\sigma_v}^2\right)$ and adding it to the original coordinates of $x$:
\begin{equation}
  x^{\prime}_{v}=x_{v} + z_v
\end{equation}
where the two coordinates of $x_{v}$ are noised separately, and the two are independent. We repeat this process $S$ times to generate $S$ 2D input samples that will be fed into an SH-HPE model to obtain multi-hypothesis predictions.

To further explain the rationality of our approach and explore the architecture, we provide the formula derivation of PRPose probabilistic modeling and the design and experiment of other paradigms in the supplementary material.

\begin{table}[t]
\setlength{\tabcolsep}{4pt}
\small
\centering
\vspace*{-1em}
\caption{Comparison of pose estimation results based on sequences ($T=243$) on the Human3.6M dataset. The P-Best and J-Best settings follow D3DP \cite{Shan_2023_ICCV}.}
\resizebox{\linewidth}{!}{
\begin{tabular}{lc|cc}
\hline
\multicolumn{2}{c|}{Method} & P1$\downarrow$ & P2$\downarrow$ \\ \hline
D3DP \cite{Shan_2023_ICCV} ($S=20, P$-$Best$)           &ICCV'23           &   39.5           &   31.2            \\ 
D3DP \cite{Shan_2023_ICCV} ($S=20, J$-$Best$)           &ICCV'23           &   35.4           &   28.7            \\  \hline
\textbf{Ours-Seq} ($S=20, P$-$Best$)               &           & \textbf{36.7}             & \textbf{29.7}  \\ 
\textbf{Ours-Seq} ($S=20, J$-$Best$)               &           & \textbf{25.5}            & \textbf{20.3}  \\ \hline
\end{tabular}}
\vspace*{-1em}
\label{tab4}
\end{table}

\section{Experiments}

\subsection{Datasets and Evaluation Metrics}

\textbf{Human3.6M} is currently the most widely used dataset for 3D human pose estimation \cite{ionescu2013human3}. We evaluate our method's performance on Human3.6M using two standard protocols, Protocol \#1 (P1) and Protocol \#2 (P2), similar to prior studies \cite{ci2023gfpose}. Samples of five subjects (S1, S5, S6, S7, and S8) are used to train the model, and samples of two subjects (S9 and S11) are used for testing.

\noindent\textbf{MPI-INF-3DHP} is a 3D human pose dataset that contains both indoor and complex outdoor scenes \cite{mehta2017monocular}. This dataset is commonly employed to evaluate a model's generalization capability. Therefore, we train our model on Human3.6M and evaluate its performance on MPI-INF-3DHP. Following prior research \cite{ci2023gfpose}, we employ the metric of Percentage of Correct Keypoints (PCK) with a threshold of 150mm to evaluate the accuracy of the 3D human pose estimation.

\begin{table}[tpb]
\setlength{\tabcolsep}{2pt}

\small
\caption{Pose estimation results on the MPI-INF-3DHP dataset. Following prior research \cite{ci2023gfpose}, $200$ samples are drawn.“GS” represents the “Green Screen”.}
\resizebox{\linewidth}{!}{
\begin{tabular}{lc|ccc|c}
\hline
\multicolumn{2}{c|}{Method}             & GS$\uparrow$            & noGS$\uparrow$           & Outdoor$\uparrow$        & ALL PCK$\uparrow$        \\ \hline
MDN \cite{li2019generating}  &CVPR'19 & 70.1           & 68.2           & 66.6           & 67.9           \\
NF \cite{wehrbein2021probabilistic}  &ICCV'21  & 86.6           & 82.8           & 82.5           & 84.3           \\
GAN \cite{li2020weakly} &BMVC'20 & 86.9           & 86.6           & 79.3           & 85.0           \\
GFPose \cite{ci2023gfpose} &CVPR'23 & 88.4           & 87.1           & 84.3           & 86.9           \\ \hline
\textbf{Ours}  &              & \textbf{93.1} & \textbf{87.6} & \textbf{89.3} & \textbf{90.2} \\ \hline
\end{tabular}
}
\vspace*{-1em}
\label{tab2}
\end{table}

\subsection{Implementation Details}

For the experiments conducted on the Human3.6M dataset, we use the CPN \cite{chen2018cascaded} for 2D pose detection, following previous single-hypothesis works \cite{cai2023htnet}. For generalization testing on the MPI-INF-3DHP dataset, we employ the pre-trained HRNet \cite{sun2019deep} as our 2D pose detector and then map to the skeletal definition of Human3.6M through interpolation.We apply the HTNet as the SH-HPE module and set the $AVG$ as an MGCN framework with a channel dimension of 256. To align with existing multi-hypothesis works, our models have not utilized the Refinement Module\cite{cai2019exploiting} often mentioned in single-hypothesis works. The parameters of the single-hypothesis model are frozen, and only the parameters of the $AVG$ model are trained. All experiments are conducted on a single NVIDIA RTX 3090 GPU. During the testing process, the coefficient $\alpha$ for adjusting the noise distribution is set to 0.005 for the Human3.6M dataset and 0.01 for the MPI-INF-3DHP dataset.



\subsection{Quantitative Results}
\textbf{Effectiveness}. We validated the effectiveness of our method on the Human3.6M dataset. Following previous works \cite{wehrbein2021probabilistic, ci2023gfpose}, we generated $S$ 3D pose estimates for each 2D pose and reported the minimum MPJPE (Mean Per Joint Position Error) between the ground truth and all estimates. As shown in Table \ref{tab1}, with 10 hypotheses, our method outperforms the GFPose \cite{ci2023gfpose} accuracy in the Protocol \#1 testing setting. With 200 hypotheses, our method is slightly lower or comparable to the state-of-the-art accuracy \cite{ci2023gfpose} in Protocol \#1 and Protocol \#2 testing settings. Where S=200 focuses on evaluating the upper limit of model performance.
In practice, given the balance between efficiency and performance, the hypothetical number is usually a modest number, such as S=10. Considering the computational efficiency advantage (improved by two orders of magnitude compared to GFPose), the superiority of our method is still significant.

In addition, due to the universality of our proposed framework, we also extend PRPose to sequence-based 2D-to-3D lifting models (named \textit{Ours-Seq}), and the results are also presented in Table \ref{tab4}. Specifically, we use MixSTE \cite{zhang2022mixste} as the SH-HPE for experimentation, with video as a sample dimension, meaning that all frames in a video have the same noise added, thereby expanding the model to multi-hypothesis generation without changing the smoothness characteristics of the video. Finally, our method significantly exceeds the latest sequence based work D3DP \cite{Shan_2023_ICCV}, further verifying the universality and enormous potential of our architecture. For more details and discussion, please refer to \textit{Section S4.1} of the supplementary material.

\begin{table}[t]
\setlength{\tabcolsep}{1.6pt}
\small
\centering
\caption{Comparison of FPS results on the Human3.6M dataset. All data were tested on the same RTX 3090, with one frame treated as one sample and multiple hypotheses generated in parallel within a single sample.}
\resizebox{\linewidth}{!}{
\begin{tabular}{l|cccc}
\hline
\multirow{2}{*}{Method}            & FPS  & FPS      & Param  & Protocol \#1 \\ 
&  ($S=10$) &  ($S=200$)      & (M)  & ($S=10/200$) \\ \hline
GraphMDN \cite{oikarinen2021graphmdn} & ---       & 121.87          & 0.69               & ---/46.2     \\
NF \cite{wehrbein2021probabilistic}  & ---       & 188.94          & 2.16               & ---/44.3     \\
\textbf{Ours-L}            & ---       & \textbf{201.51} & \textbf{0.17} & ---/\textbf{42.9}    \\ \hline
GFPose \cite{ci2023gfpose} & 0.56      & 0.54            & 12.50              & 45.1/\textbf{35.6}    \\ 
\textbf{Ours-B}            & \textbf{66.62}     & \textbf{59.29}           &\textbf{4.14}          & \textbf{44.1}/38.2  \\ \hline
\end{tabular}}
\vspace*{-1em}
\label{tab3}
\end{table}

\begin{table}[t]
\setlength{\tabcolsep}{3pt}
\small
\centering
\caption{Ablation experiment of adaptive noise on MPI-INF-3DHP dataset. "No adapted" denotes constant variance without adjustment. "Sample-Joints adapted" follow the main experiment settings.}

\resizebox{\linewidth}{!}{
\begin{tabular}{l|cc}
\hline
       Strategy          & PCK ($S=10$)   & PCK ($S=200$)   \\ \hline
No adapted                   & 82.47          & 86.11           \\
Joints adapted             & 82.57          & 86.17           \\
Sample adapted                & 84.40          & 89.12           \\
Sample-Joints adapted       & \textbf{84.61}          & \textbf{90.19}            \\ \hline
\end{tabular}}
\vspace*{-1em}

\label{tab5}
\end{table}

\textbf{Generalization}. We evaluated the generalization ability of our model on the MPI-INF-3DHP dataset. All parameters of the model were trained on the Human3.6M dataset and were not fine-tuned on the MPI-INF-3DHP dataset. As shown in Table \ref{tab2}, our method outperformed the state-of-the-art approaches \cite{wehrbein2021probabilistic}, \cite{ci2023gfpose}, and \cite{li2020weakly} by a significant margin. Even compared with two models that use ground-truth 2D pose \cite{li2019generating} and are specifically designed for transfer learning \cite{li2020weakly}, our method still achieved a remarkable improvement. Unlike other methods, our approach did not suffer a decrease in accuracy compared to the "noGS" condition in complex outdoor environments. This is due to the regulation of the adaptive noise generation module, which allowed the correction of joints with large estimation errors.

\textbf{Efficiency}. To prove the advantages of our model on computational efficiency, we compare the frames per second (FPS) of our model with the state-of-the-art methods. In addition to our basic model defined above (hereby named \textit{Ours-B} for distinguish), we propose a lightweight mode (named \textit{Ours-L}), where the SH-HPE module is a lightweight version of MGCN \cite{zou2021modulated} with a channel dimension of 128 that removes non-local modules and shares the same weight among different nodes, and $AVG$ is implemented with a single-block Simple-Baseline3D \cite{martinez2017simple}.  

The performance and efficiency comparison is shown in Table \ref{tab3}. The results illustrate that \textit{Ours-B} achieved advanced estimation accuracy with a high inference speed. Especially, compared with GFPose, the state-of-the-art method of MH-HPE, \textit{Ours-B} obtained more than 100 times speed gain at a comparable accuracy. Besides, \textit{Ours-L} achieved the highest FPS while maintaining a competitive accuracy. The results indicate that our method can achieve a perfect balance between speed and accuracy.

\textbf{Ablation Study for Adaptive Noise}. We conducted ablation experiments to observe the impact of the adaptive noise generation module on the model's accuracy and generalization ability. The experimental results are shown in Table \ref{tab5}. It can be found that compared with no adapted or single adapted, Sample-Joints adapted shows better performance, indicating the necessity of adaptive noise modules. With the increase in the number of hypotheses, the probability of sampling the expected sample increases, and the advantages of adaptive noise become more obvious.

\begin{figure}[t]
  \centering
  \includegraphics[width=1\linewidth]{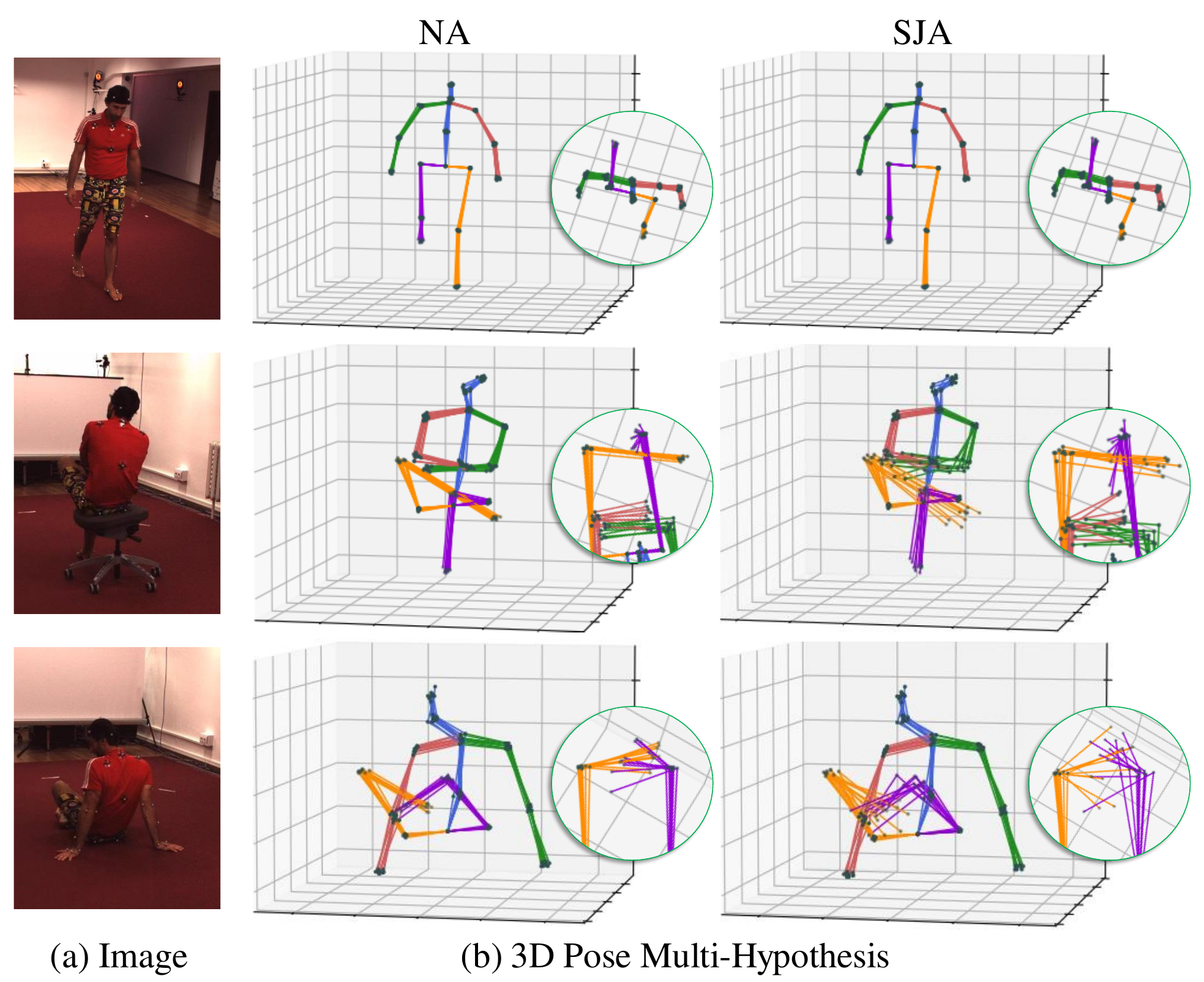}
  \caption{Qualitative comparison of multi-hypothesis outputs for Sample-Joints adapted (SJA) and No adapted (NA) methods on the Human3.6M dataset.}
  \label{fig4}
  \vspace*{-1em}
\end{figure}

\subsection{Qualitative Results}

To further demonstrate the effectiveness of our framework and adaptive noise addition strategy, we visualized module outputs and compared with No adapted strategy. Figure \ref{fig4} displays results for Waiting, Smoking, Sitting Down actions. (During Smoking and Sitting Down action, some joints are obviously blocked.) Compared with No adapted strategy, our method can generate more diverse 3d human poses at joints with large estimation error and high depth uncertainty. This further suggests that adaptive noise is more effective in difficult samples. For those simple samples, fixed distribution noise can already complete the bias correction, making the advantage of adaptive noise less obvious. When the sample quality is low, fixed distribution noise is difficult to apply to various situations, and the advantages of adaptive noise are manifested.


\section{Conclusion}

Aiming at the ill-posed problem of 2D-to-3D lifting in 3D HPE and the high computational cost of existing multi-hypothesis methods, we propose a general framework called PRPose for the probabilistic modeling process reconstruction of 3D human pose estimation. With adaptive noise sampling, it seamlessly extends lightweight single-hypothesis models to the multi-hypothesis method, significantly enhancing computational efficiency while maintaining state-of-the-art accuracy. Through extensive experiments, we demonstrate the effectiveness and rationality of our approach. Our work provides a new way of thinking for 3D HPE, enabling multi-hypothesis research to directly benefit from improving single-hypothesis models. Our work also reveals the great significance of researching adaptive noise generation strategies based on human topology and 2D pose, and we will explore these issues in the future.



\bibliographystyle{IEEEtran}
\bibliography{conference_101719}
\newpage
\appendix


In the supplementary materials, we provide additional insights and details regarding our novel approach, PRPose, for the Probabilistic Restoration of SH-HPE in the 2D-to-3D lifting process. These materials include comprehensive probabilistic modeling derivations, extensions of our methods and further exploration of different paradigms, and more relevant experiments and analyses.

\subsection{Probability modeling derivation of PRPose}

In this section, we present a comprehensive probabilistic modeling derivation of our novel approach, PRPose. Our method focuses on capturing the underlying probability distribution inherent in the 2D-3D lifting process of the SH-HPE model. This is achieved by introducing adaptive noise mapping to the 2D pose inputs, effectively reconstructing the original probabilistic modeling process of the SH-HPE method. 

\noindent 
\textit{\textbf{A.1 Formulation of Optimization Objective}}

Next, the methodology is outlined within the context of probability modeling. Given an input 2D pose $\mathbf{x} \in \mathbb{R}^{2V}$, where $V$ represents the number of joints, and a corresponding ground truth 3D pose $\mathbf{y} \in \mathbb{R}^{3V}$, our primary objective is to establish a likelihood-based framework that effectively aligns the probability distribution of $\mathbf{y}$ with the input $\mathbf{x}$.
Mathematically, we seek to maximize the logarithmic likelihood function given $\mathbf{x}$, expressed as:
\begin{equation}
    L=\sum_{\mathbf{x}} \log{\mathcal{P} (\mathbf{y} \mid \mathbf{x})} 
\end{equation}

figure 1.c illustrates the process of PRPose's probability modeling. For clarity, we use $\mathbf{z}$ to replace $\mathbf{x'}$ in the main text, serving as an intermediate distribution's hidden variable. Specifically, $\mathbf{z} \in \mathbb{R}^{2V}$ is a hypothesis for the 2D pose, incorporating the influence of adaptive noise. $\mathcal{P}(\mathbf{z} \mid \mathbf{x})$ represents the hidden probability distribution that governs this mapping; in other words, it denotes the distribution of $\mathbf{z}$ given the input $\mathbf{x}$.

Our objective is to estimate this distribution $\mathcal{P}(\mathbf{z} \mid \mathbf{x})$ by introducing adaptive noise, thus capturing the inherent uncertainty in the process. To achieve this, we introduce an approximate distribution $\mathcal{Q}(\mathbf{z} \mid \mathbf{x})$, which is a model-predicted distribution used to approximate the true posterior distribution. By continuously refining $\mathcal{Q}(\mathbf{z} \mid \mathbf{x})$, we aim to better capture the uncertainty in the pose transformation process dictated by the hidden probability distribution $\mathcal{P}(\mathbf{z} \mid \mathbf{x})$. 

\noindent 
\textit{\textbf{A.2 Derivation of Probabilistic Modeling}}

Next, we will further derive $\log \mathcal{P}(\mathbf{y} \mid\mathbf{x})$, which is the logarithmic probability of the output $\mathbf{y}$ given the input $\mathbf{x}$. This derivation will unveil how we employ the introduced approximate distributions to efficiently approximate this probability and how these distributions are combined to form the probabilistic modeling process of PRPose.

\begin{equation}
\begin{aligned}
\log \mathcal{P}(\mathbf{y} \mid \mathbf{x}) & = \int_{\mathbf{z}} \mathcal{Q}(\mathbf{z} \mid \mathbf{x}) \log \mathcal{P}(\mathbf{y} \mid \mathbf{x}) d\mathbf{z} \\
& = \int_{\mathbf{z}} \mathcal{Q}(\mathbf{z} \mid \mathbf{x}) \log \left(\frac{\mathcal{P}(\mathbf{y}, \mathbf{z}, \mathbf{x})}{\mathcal{P}(\mathbf{z} \mid \mathbf{x}, \mathbf{y}) \mathcal{P}(\mathbf{x})}\right) d\mathbf{z} \\
& = \int_{\mathbf{z}} \mathcal{Q}(\mathbf{z} \mid \mathbf{x}) \log \left(\frac{\mathcal{P}(\mathbf{y} \mid \mathbf{z}, \mathbf{x}) \mathcal{P}(\mathbf{z} \mid \mathbf{x})}{\mathcal{P}(\mathbf{z} \mid \mathbf{x}, \mathbf{y}) }\right) d\mathbf{z} \\
& \begin{aligned}
= & \int_{\mathbf{z}} \mathcal{Q}(\mathbf{z} \mid \mathbf{x}) \log \left(\frac{\mathcal{Q}(\mathbf{z} \mid \mathbf{x})}{\mathcal{P}(\mathbf{z} \mid \mathbf{x}, \mathbf{y}) }\right) d\mathbf{z} +\\
& \int_{\mathbf{z}} \mathcal{Q}(\mathbf{z} \mid \mathbf{x}) \log \left(\frac{\mathcal{P}(\mathbf{z} \mid \mathbf{x})}{\mathcal{Q}(\mathbf{z} \mid \mathbf{x}) }\right) d\mathbf{z} +\\
& \int_{\mathbf{z}} \mathcal{Q}(\mathbf{z} \mid \mathbf{x}) \log \left(\mathcal{P}(\mathbf{y} \mid \mathbf{z}, \mathbf{x}) \right) d\mathbf{z} \\
= & KL(\mathcal{Q}(\mathbf{z} \mid \mathbf{x}) \mid\mid \mathcal{P}(\mathbf{z} \mid \mathbf{x}, \mathbf{y})) - \\
& KL(\mathcal{Q}(\mathbf{z} \mid \mathbf{x}) \mid\mid \mathcal{P}(\mathbf{z} \mid \mathbf{x})) + \\
& \mathbb{E}_{\mathcal{Q}}[\log \left(\mathcal{P}(\mathbf{y} \mid \mathbf{z}, \mathbf{x}) \right)]
\end{aligned}
\end{aligned}
\end{equation}

The Evidence Lower Bound (ELBO) is a fundamental concept in variational inference \cite{Blei_2017}. It provides a lower bound on the log-likelihood of observed data, which can be maximized to optimize the parameters of a probabilistic model. According to the ELBO theorem, the optimization objective can be further simplified as follows:

\begin{equation}
\begin{aligned}
ELBO= & - KL(\mathcal{Q}(\mathbf{z} \mid \mathbf{x}) \mid\mid \mathcal{P}(\mathbf{z} \mid \mathbf{x})) + \\
& \mathbb{E}_{\mathcal{Q}}[\log \left(\mathcal{P}(\mathbf{y} \mid \mathbf{z}, \mathbf{x}) \right)]
\end{aligned}
\end{equation}
\noindent 
\textit{\textbf{A.3 Derivation of Loss Function}}

The objective of the first term is to minimize the KL divergence between the model output distribution $\mathcal{Q}$ and the prior distribution $\mathcal{P}$. Considering the varying noise levels across different joints in observed human poses across different samples, we have an adaptive design for the prior distribution of each joint. This enables our model to dynamically adjust and respond to diverse input scenarios. Specifically, on the one hand, we utilize the input 2D poses as the mean of the distribution. On the other hand, we observe that the distances between the 3D joint coordinates estimated by the SH-HPE model and the ground truth 3D coordinates can be approximately back-propagated in proportion to the different 2D joint inputs. Hence, we normalize the distances and utilize them as variances $\hat{A}^{\sigma}$ for the corresponding joint distributions (referred to as pseudo-labels in the main text). It's important to note that since we're considering distance errors in the joint dimensions, the variances across different dimensions within a joint are treated as equal. The mathematical formulas are expressed as follows:
\begin{equation}
A^{\sigma}=A V G_{\varphi}(\mathbf{x})
\end{equation}
\begin{equation}
\mathcal{Q}({z_v} \mid \mathbf{x})=\mathcal{N}(x_v,\sigma_v^2)
\end{equation}
\begin{equation}
\mathcal{P}({z_v} \mid \mathbf{x})=\mathcal{N}(x_v,\hat{\sigma}_v^2)
\end{equation}
where $A^{\sigma} = \{\sigma_1,...,\sigma_V\}$ is the set of output adaptive variance and $\hat{A}^{\sigma} = \{\hat{\sigma}_1,...,\hat{\sigma}_V\}$ is the set of pseudo-labels variance. ${z_v}$ and ${x_v}$ respectively represent variables on the v-th joint of the intermediate 2D pose hypothesis $\mathbf{z}$ and the input 2D pose $\mathbf{x}$. Further, use two distributions to calculate KL divergence:

\begin{equation}
    KL(\mathcal{Q}({z_v} \mid \mathbf{x}) \mid\mid \mathcal{P}({z_v} \mid \mathbf{x}))=\frac{1}{2} *[\log(\frac{\hat{\sigma}_v }{\sigma_v} )^2+\frac{{\sigma_v}^2}{{\hat{\sigma}_v }^2}-1 ]
\end{equation}

According to the right term function properties, this term can be further simplified to minimize the squared difference between the predicted variance and the prior variance:

\begin{equation}
  \mathcal{L}_v =\left(\sigma_v-\hat{\sigma}_v\right)^{2}
\end{equation}

The objective of the second term involves maximizing the approximate expectation between predicted 3D poses and ground truth 3D poses. This entails enhancing the value of $\mathcal{P}(\mathbf{z} \mid \mathbf{x}, \mathbf{y})$ based on the given distribution $\mathcal{Q}(\mathbf{z} \mid \mathbf{x})$. In our research, $\mathcal{P}(\mathbf{z} \mid \mathbf{x}, \mathbf{y})$ is facilitated using a pre-trained SH-HPE model, which is also optimized through maximum expectation. This optimization enables the model to generate relatively reasonable 3D poses when provided with 2D poses augmented with additional adaptive noise.

In our process of probabilistic modeling reconstruction, our objective is to train the AVG model exclusively to generate adaptive noise that accommodates different samples. We find the necessity of the second term to be dispensable. On the one hand, the parameters of the SH-HPE model remain fixed, eliminating the need for backpropagation training through maximizing the approximate expectation between predicted 3D poses and ground truth 3D poses. On the other hand, optimizing the first term is achieved by leveraging the pseudo-labels derived from the discrepancies between predicted 3D poses and ground truth 3D poses. This injects the maximum expectation into the optimization of the first term. In other words, the impact of the second term's maximum expectation on the AVG model training aligns with minimizing the KL divergence between distributions $\mathcal{Q}$ and $\mathcal{P}$. Furthermore, the input dependency of the SH-HPE model within the maximum expectation term due to multiple samplings from distribution $\mathcal{Q}$ leads to slow convergence during model training. Hence, to reduce computational complexity during training and expedite convergence, we further filter the maximum expectation term.

Therefore, the ultimate optimization objective of the maximum logarithm likelihood $\sum_{\mathbf{x}} \log{\mathcal{P} (\mathbf{y} \mid \mathbf{x})}$ is equivalent to minimizing the following loss function:
\begin{equation}
  \mathcal{L}=\sum_{\mathbf{x}} \frac{\sum_{v \in V}\left(\sigma_v-\hat{\sigma}_v\right)^{2}}{V}
\end{equation}


\subsection{Architecture Exploration}

In this section, we explore the design of different paradigms of PRPose architecture based on the existing SH-HPE model. Then the advantages and applicable scenarios of these two paradigms are compared and discussed through experiments.
\noindent 
\textit{\textbf{B.1 Collaborative Paradigms}}

\begin{figure}[tpb]
  \centering
  \includegraphics[width=\linewidth]{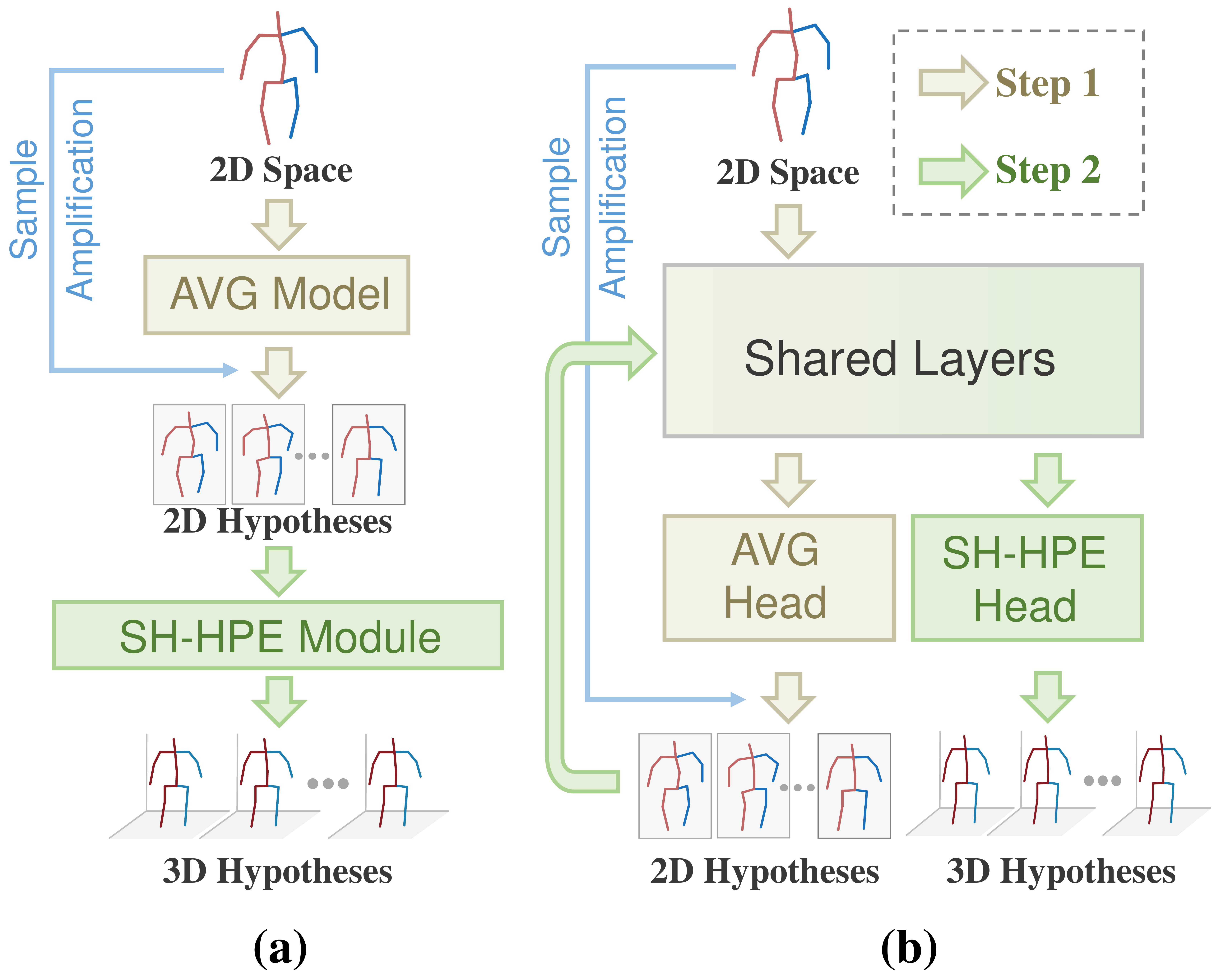}
  \caption{Two paradigms of $AVG$ embedded into the PRPose framework: (a) the GCN layer comprising the $AVG$ has an independent weight; (b) $AVG$ shares part of the GCN weight with the SH-HPE model, and only one additional mapping head is needed to obtain the noise estimation.}
  \label{sfig3}
\end{figure}
The input of $AVG$ is a 2D skeleton, and the output is an adaptive variance matrix, so we use the Graph Convolutional Neural network and fully connected layer to construct $AVG$. Considering that both $AVG$ and the SH-HPE model need to encode 2D skeletons, we design two collaborative paradigms of $AVG$ and the SH-HPE model:

\begin{itemize}
 \item The independent paradigm is shown in Figure \ref{sfig3}.a, in which $AVG$ and the SH-HPE model are independent, and the parameters of the two do not affect each other. This paradigm needs to design $AVG$ separately, increasing the number of model parameters to a certain extent.

 \item The shared paradigm is shown in Figure \ref{sfig3}.b. The initial layers of the neural network has been proven to learn to encode common features \cite{yosinski2014transferable}, which inspires us to share the parameters between the $AVG$ and the shallow graph convolutional neural network of the single-hypothesis model, and only need to add an additional prediction head to complete the noise estimation. This paradigm hardly increases any parameters, but $AVG$ relies heavily on the single-hypothesis model, which loses flexibility in design and increases the amount of computation in the inference stage.
\end{itemize}

The pose estimation accuracy of the two collaborative paradigms is similar, but they have their own advantages and disadvantages. The independent paradigm has higher flexibility and less inference calculation, and the shared paradigm has fewer parameters. We can flexibly choose according to the actual application requirements. Please refer to the experimental part for specific results and analysis.

\begin{table}[t]
\setlength{\tabcolsep}{2pt}
\small
\centering
\caption{Performance comparison of different paradigms on the Human3.6M dataset.}
\resizebox{\linewidth}{!}{
\begin{tabular}{cccccc}
\hline
Paradigm                         & Param(M)                        & Num of hypotheses & P1    & P2    & FPS            \\ \hline
\multirow{2}{*}{\begin{tabular}[c]{@{}c@{}}Independent\\ Paradigm\end{tabular}} & \multirow{2}{*}{4.14} & S=10                 & 44.08 & 35.45 & \textbf{66.62}          \\  
                         &                                     & S=200                & 38.18 & 30.98 & \textbf{59.29} \\ \hline
\multirow{2}{*}{\begin{tabular}[c]{@{}c@{}}Shared\\ Paradigm\end{tabular}} & \multirow{2}{*}{\textbf{3.10}} & S=10                 & 44.13 & 35.46 & 53.98          \\  
                                     &                                     & S=200                & 38.14 & 30.90 & 47.66          \\ \hline
\end{tabular}}
\label{stab5}
\end{table}
\noindent 
\textit{\textbf{B.2 Experiments of Collaborative Paradigms}}

Table \ref{stab5} presents performance comparisons between the two collaborative paradigms for the $AVG$ and single-hypothesis model. The independent paradigm enables separate $AVG$ design for flexibility and efficiency, while the shared paradigm offers parameter efficiency. Both paradigms exhibit comparable accuracy, allowing selection based on specific needs. The independent paradigm suits high real-time demands, while the shared paradigm suits parameter-constrained scenarios.

\subsection{More Experiments}

\noindent 
\textit{\textbf{C.1 Extension of PRPose to sequence-based HPE}}

In the experiment (\textit{Section 3.3}) of the main text, We also extend PRPose to sequence-based 2D-to-3D lifting models \cite{pavllo20193d, zheng20213d, zhang2022uncertainty} to further verify the universality and powerful potential of our proposed framework, and the results are also given in Table \ref{tab4}. However, due to space constraints in the main text, we will discuss and provide additional details here.



Regarding implementation details of the AVG model in the extension of PRPose to sequence-based HPE, our main objective was to validate the versatility of the architecture, so we utilized existing models as the backbone network without extensive revision. Specifically, we utilized the architecture shown in Figure \ref{fig2}. Considering that the extension to video is for sequence-based pose inputs, the AVG model incorporates a lightweight version of MixSTE, where some specific parameter adjustments are made, including setting the intermediate dimension to 256, the depth to 4, and the number of attention heads to 4. Finally, pooling operations were performed along the frame dimension to generate the output. In addition, because all frames in a video dimension use the same noise, in order to increase the diversity of adaptive noise input, we randomly mask the noise of some frames with a probability of 0.2, that is, the masked frames do not add adaptive noise.

In the above, we have explored the application of adaptive noise fitting at the video level, where the same noise is added to all frames of the same video, while still retaining the adaptive ability of different joints between different samples (in the context of sequence-based 3D pose estimation, the sample dimension represents the video). In addition, we also explored the application of adaptive noise fitting during training at the frame level within videos. However, this strategy demonstrated slightly less favorable outcomes in comparison to the strategy of adding noise at the video level. Our initial analysis suggests that adapting at the frame level could potentially impact the smoothness and other inherent characteristics of the video. In our future research endeavors, we aim to investigate techniques for integrating adaptive noise capabilities into video-based 3D human pose estimation while safeguarding the intrinsic smoothness and other essential attributes of the video.


\begin{table}[]
\setlength{\tabcolsep}{1.5pt}
\small
\centering
\caption{Ablation study of noise sampling positions on the Human3.6M dataset. "Pre-sample" refers to 2D sample amplification, and "Post-sample" refers to 3D sample amplification.}
\resizebox{\linewidth}{!}{
\begin{tabular}{ccccc}
\hline
Strategies  & P1($S=10$) & P2($S=10$) & P1($S=200$) & P2($S=200$) \\ \hline
Pre-sample  & 44.08    & 35.45    & 38.18     & 30.98     \\ \hline
Post-sample & 47.62    & 38.03    & 43.71     & 34.79     \\ \hline
\end{tabular}}
\vspace*{-1em}
\label{stab7}
\end{table}

\begin{figure}[t]
  \centering
  \includegraphics[width=1\linewidth]{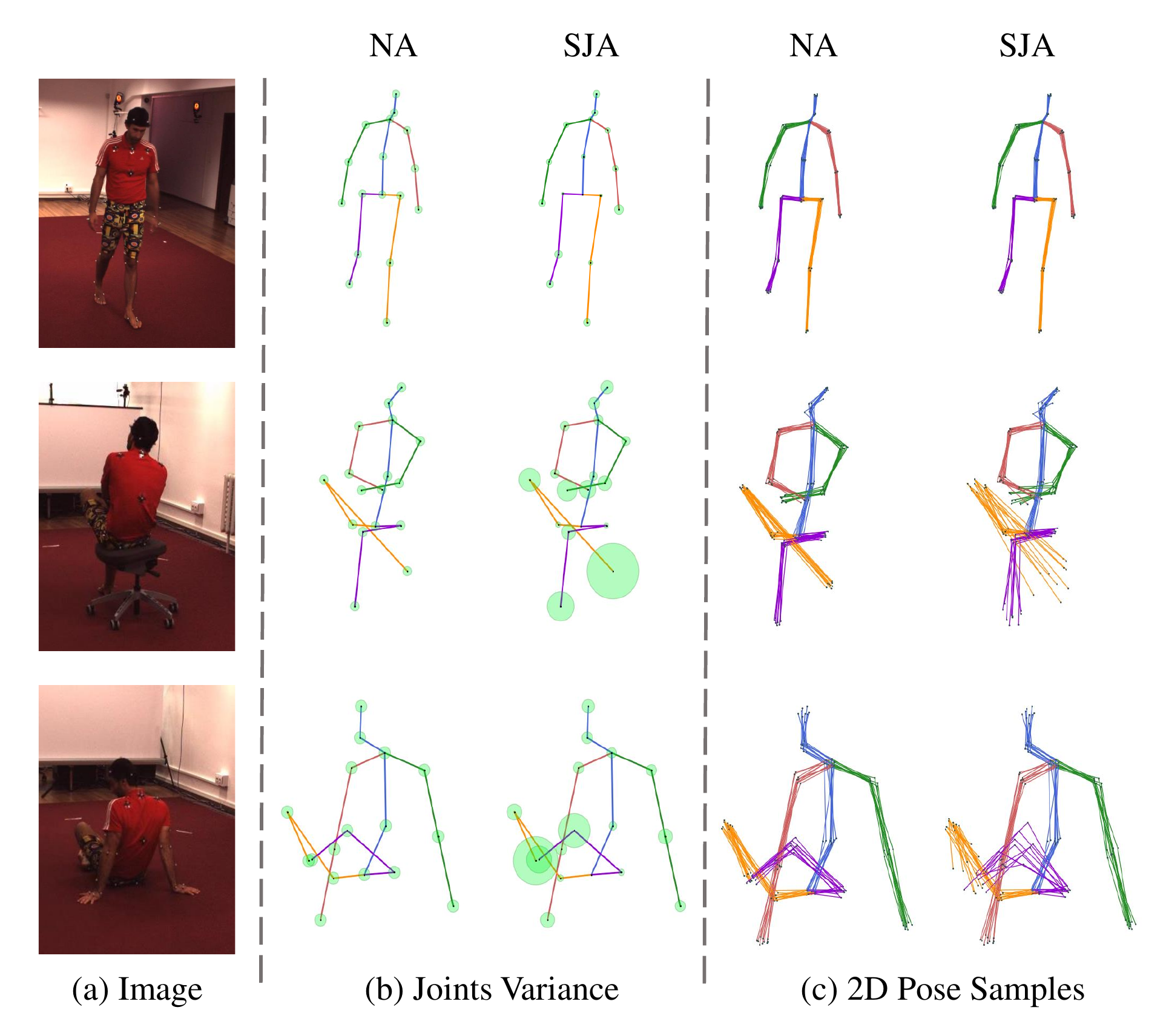}
  \caption{Qualitative comparison of the outputs at different intermediate stages for Sample-Joints adapted (SJA) and No adapted (NA) methods on the Human3.6M dataset. From left to right: (a) the original image, (b) the results of applying a constant variance/variance matrix generated by $AVG$ to each joint, where the diameter of the circle indicates the magnitude of the variance, (c) the 2D samples amplificated by adaptive noise sampling. }
  \label{sfig4}
\end{figure}

\noindent 
\textit{\textbf{C.2 Layer of Adding Noise}}

In the previous experiments, we performed 2D sample amplification by adding noise at the input (2D Pose) of the single-hypothesis model. In fact, we can also directly add noise to the output (3D Pose) of the single-hypothesis model to form a multi-hypothesis estimate. The experimental results are shown in Table \ref{stab7}. It can be seen that the post-sampling method (i.e. adding noise in 3D space) lags behind the pre-sampling method (adding noise in 2D space) in various indicators. We speculate that: 1) The Pre-sample is conducted in 2D space, while the Post-sample is conducted in 3D space, which results in a much larger sampling space for the Post-sample and a lower probability of sampling the plausible hypotheses; 2) after the Pre-sample, the samples will be processed by the single hypothesis model, and some unreasonable samples have the opportunity to be adjusted back to a normal level, while after the Post-sample, the samples are directly outputted without further adjustment.

\noindent 
\textit{\textbf{C.3 More Qualitative Results}}

In the qualitative results (\textit{Section 3.5}) of the main text, we have compared and analyzed the multi-hypothesis outputs of Sample-Joints adapted (SJA) and No adapted (NA) methods by visual experiments. Here we further show the visual experimental comparison of the output in the intermediate stage, as shown in Figure \ref{sfig4}.

Observing the second column of Figure \ref{sfig4}, it can be noticed that for different actions, the variance of each joint under the No adapted strategy is the same, despite some joints having high uncertainty. In contrast, the variance matrix generated by $AVG$ has sample-joints adaptivity. For the Waiting action, where there is no obstruction in the image and the limbs are extended, the 2D estimation is relatively accurate, indicating that the adaptive noise variance is small, and therefore, a large range of random sampling is not required. For the Smoking action, where there is a large area of obstruction in the image, the legs and hands pose can only be inferred by common sense, and the 2D estimation of these parts is highly inaccurate, resulting in a larger adaptive variance for these parts in this sample. For the Sitting Down action, only the feet are obstructed, and correspondingly, the adaptive variance of the feet is larger than that of other joints. These results demonstrate the effectiveness of $AVG$.

The third column in Figure \ref{sfig4} represent multiple 2D samples obtained through noise sampling. It can be seen that after adaptive noise sampling, the difficult samples obtained more diverse 2D amplifications, while the amplifications obtained from the No adapted noise sampling are relatively conservative. Accordingly, SJA is able to generate more diverse 3D poses at nodes with large estimation errors and high depth uncertainty (as shown in Figure \ref{fig4} in the main text).
\end{document}